\documentclass[11pt]{article}

\usepackage[final]{acl}
\usepackage{times}
\usepackage{latexsym}
\usepackage{float}
\usepackage[normalem]{ulem}

\usepackage[T1]{fontenc}
\usepackage[utf8]{inputenc}

\usepackage{tipa}               % REMOVE TODO NOTES before send
\usepackage{orcidlink}
\usepackage{todonotes}          % REMOVE TODO NOTES before send
\usepackage{rotating}
\usepackage{tabularray}
\usepackage{booktabs}
\usepackage{microtype}

\usepackage{inconsolata}

\usepackage{graphicx}

\title{Building Community-Centred NLP Resources for Puno Quechua}

\author{
 \textbf{Elwin Huaman\textsuperscript{1}\orcidlink{0000-0002-2410-4977}},
 \textbf{Adrian Gamarra Lafuente\textsuperscript{2}\orcidlink{0009-0006-6829-9741}},
 \textbf{Johanna Cordova\textsuperscript{3}\orcidlink{0000-0001-5950-5271}},
 \textbf{Anna Korhonen\textsuperscript{1}\orcidlink{0000-0002-3692-3144}}
\\
\\
 \textsuperscript{1}University of Cambridge (UK),
 \textsuperscript{2}Stanford University (USA),
 \textsuperscript{3}ERTIM - Inalco (France)
\\
 \small{
   \textbf{Correspondence:} \href{mailto:elh97@cam.ac.uk}{elh97@cam.ac.uk}, 
        \href{mailto:agamarra@stanford.edu}{agamarra@stanford.edu},
        \href{mailto:johanna.cordova@inalco.fr}{johanna.cordova@inalco.fr}
 }
}
\begin{document}
\maketitle
\begin{abstract}
The preservation of under-resourced languages requires digital tools and resources shaped by and for their speakers. We present the first dedicated ASR resources for Puno Quechua (ISO 639-3: \texttt{qxp}):
(1) the largest speech corpus for any single Quechua variety, consisting in 66 hours of recordings for scripted and spontaneous speech (including 36 hours of manually transcribed and validated data), collected via a participatory design campaign;
(2) the first systematic ASR benchmark for Puno Quechua, evaluating state-of-the-art models and fine-tuning Whisper-base, wav2vec2-base, and XLS-R-300M, with and without continued pre-training (CPT);
(3) an open release of all datasets and fine-tuned models. 
\end{abstract}

\section{Introduction}
\label{sec:1-intro}

The revitalization of indigenous languages depends increasingly on digital tools that promote language use and confer economic and social value~\cite{Galla2016Indigenous}. Puno Quechua (\texttt{qxp}) is spoken by approximately 465,000 people in the Puno region of Peru\footnote{Estimation based on 2017 Peruvian National Census, \url{https://www.inei.gob.pe/media/MenuRecursivo/publicaciones_digitales/Est/Lib1563/}}, yet formal education is conducted almost exclusively in Spanish, leaving its speakers largely illiterate in their own language. This literacy gap prevents native speakers from interacting with text-input AI applications (ChatGPT or Gemini), excluding them from the growing digital ecosystem.

\begin{figure}[htbp]
\centering
  \begin{minipage}{0.48\textwidth}
    \centering
    \includegraphics[width=\textwidth]{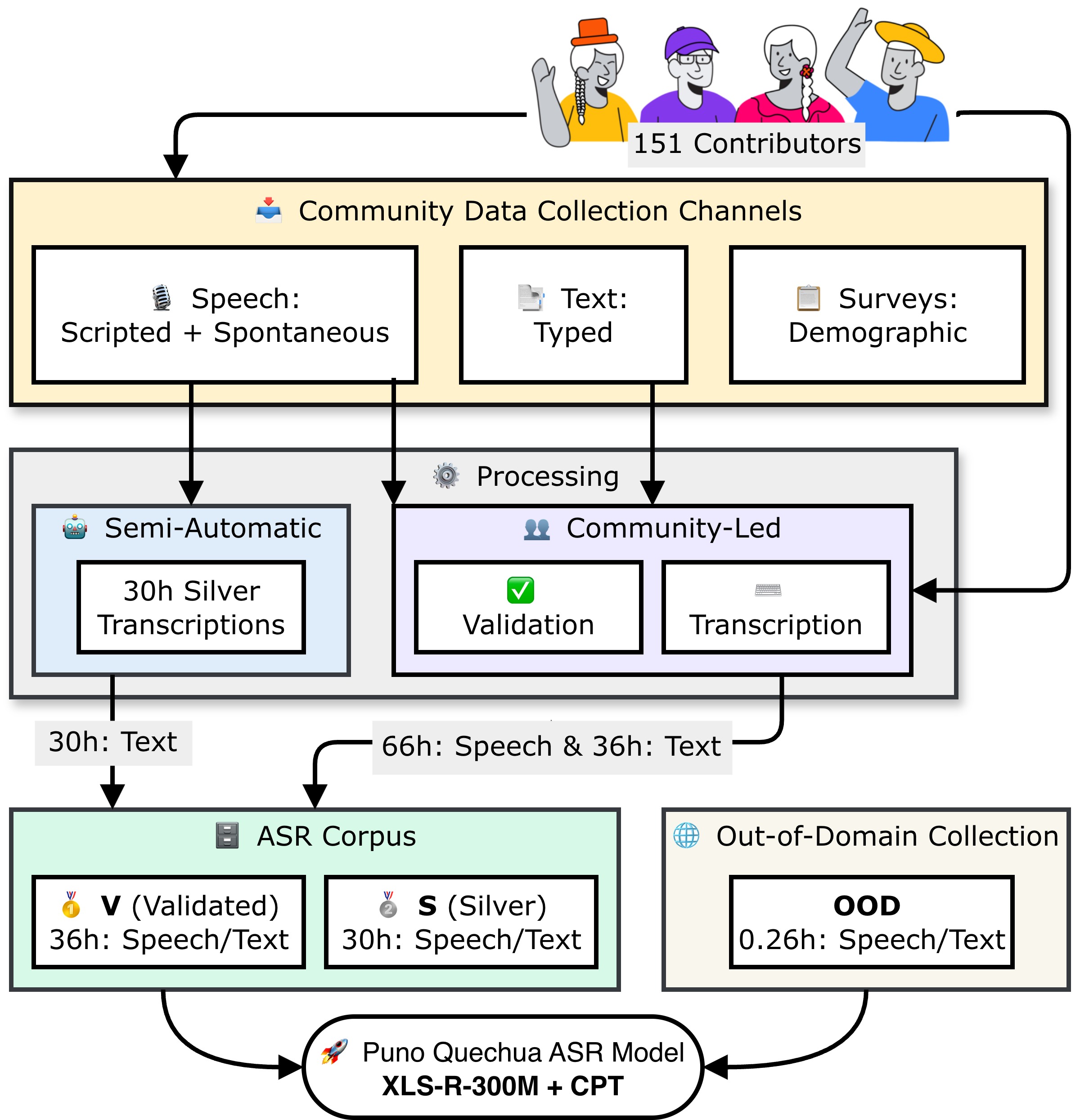}
  \end{minipage}
  \caption{Puno Quechua ASR Pipeline.}
  \label{fig:asr-qxp-pipeline}
\end{figure}

An Automatic Speech Recognition (ASR) system represents a more human-centred solution: rather than forcing community members to adapt to text-heavy interfaces, ASR can adapt technology to the oral-centric reality of communities. Despite this need, no dedicated ASR resources exist for Puno Quechua. Prior work on Quechua ASR tends to treat the language family as a homogeneous entity: the variant(s) in question are sometimes not formally identified, or they are aggregated by linguistic group (e.g. Southern Quechua \cite{cardenas18} or Collao Quechua \cite{Paccotacya22}) without any proper examination of how their differences may affect practical applications. Furthermore, existing corpora suffer from data scarcity, restricted access (\textit{Siminchik} by~\citeauthor{cardenas18}, one of the largest corpora referenced in the scientific literature, has never been published in open access), and the absence of a variety-specific benchmark.
%aggregating distinct varieties under a single label and suffers from data scarcity, restricted access, and the absence of a variety-specific benchmark.
Figure~\ref{fig:asr-qxp-pipeline} shows an overview of the Puno Quechua ASR pipeline.

This paper makes three contributions: (1) the largest ASR corpus for a Quechua variety via participatory data collection; (2) the first systematic ASR benchmark for Puno Quechua; and (3) open-sourced code and splits,\footnote{\url{https://github.com/QuechuaBase/asr-puno-quechua}} datasets,\footnote{\url{https://mozilladatacollective.com/datasets}} and models.\footnote{\url{https://huggingface.co/QuechuaBase}}

% \textcolor{red}{The paper is structured as follows. Section~\ref{sec:2-background} describes the Puno Quechua language and reviews existing ASR systems for Quechua. Section~\ref{sec:3-data-collection} details our participatory design methodology and data collection process. Afterwards, Section~\ref{sec:4-evaluation} describes the training and evaluation protocols for ASR models. Finally, we present our results in Section~\ref{sec:5-results}, followed by discussion in Section \ref{sec:6-discussion}. Last but not least, in Section~\ref{sec:7-conclusion} we highlight our conclusions and directions for future work.}

\section{Background}
\label{sec:2-background}

\paragraph{Puno Quechua (\texttt{qxp}).}
It belongs to the Southern Quechua branch (QIIC)~\cite{Torero1970} and is characterised by a rich consonant inventory of 26 phonemes, including ejective, aspirated, and uvular stops, and three vowels (/a/, /i/, /u/). It exhibits Aymara influence in its phonology, suffix inventory, and vocabulary~\cite{adelaar87}. 
% \textcolor{red}{These features, particularly the ejective and uvular consonants, are acoustically distinctive and under-represented in the pre-training data of existing multilingual speech models}.

\paragraph{ASR for Quechua (\texttt{que}).} 
ASR development for Quechua faces several compounding challenges: (a) Data scarcity: labelled speech data are extremely scarce, and when corpora exist they often aggregate varieties under a macrolanguage label (\texttt{que}), conflating varieties that are not mutually intelligible; 
(b) Morphological complexity: Quechua is highly agglutinative, resulting in poor word-level WER metrics; (c) Low written literacy: most speakers are illiterate in Quechua, making community-driven transcription difficult, 
% (d) No standard orthography adopted, competing writing systems promoted by governments, missionaries, and academics create annotation inconsistency; 
and (d) No variety-specific benchmark: to our knowledge, no prior work establishes a consistent evaluation benchmark for an individual Quechua variety. Recent work by~\citet{Keren25} covers 30+ Quechua varieties within omnilingual ASR, providing reference baselines including for \texttt{qxp}.
\section{Participatory Design Data Collection}
\label{sec:3-data-collection}

We collected the Puno Quechua speech corpus through a four-phase participatory design process~\cite{HuamanHH25,Petti26,Spinuzzi05,Wilson25}: 
\paragraph{Planning.} Identifying the ISO 639-3 code \texttt{qxp}, establishing partnerships with the National University of Altiplano Puno and the local community Illariy Ch'aska, and assessing community needs.
\paragraph{Preparation.} Setting up data governance under CC0-1.0 licence, preparing seed sentences and questions covering agriculture, healthcare, and technology, and localising the Mozilla Common Voice platform to Puno Quechua.\footnote{\url{https://commonvoice.mozilla.org/qxp}} 
\paragraph{Collection.} Voluntary, skill-based contributions such as reading, speaking, listening, or writing, with community-led validation and privacy-preserving processing data. 
\paragraph{Deployment.} Open release on Mozilla Data Collective,\footnote{\url{https://mozilladatacollective.com/}} certificates of contribution, voucher incentives for participants, and impact assessment.

The campaign ran from January to February 2026. A total of 396 volunteers registered, 292 were confirmed, 151 contributed, and 31 completed the full campaign. 
The resulting speech data have been aggregated and released as Common Voice Scripted Speech v25 (including v23 and v24 by ~\citeauthor{HuamanHHQ26}) with a total of 34.81 hours (30.5 validated) and Common Voice Spontaneous Speech v3 (including v1 and v2 by ~\citeauthor{HuamanHHQ26}) with a total of 35.3 hours (5.18 validated).

Table \ref{tab:dataset-corpus} summarizes the processed data that has been collected and can be used for training and evaluating models.

 \begin{table}[t]
    \centering
    \def\arraystretch{1.1}
    \small
    \begin{tabular}{|l|l|r|r|}
        \hline
        \textbf{Dataset} & \textbf{Type} & \textbf{Validated (}\texttt{V}\textbf{)} & \textbf{Silver (}\texttt{S}\textbf{)} \\
        \hline
        \texttt{SCS-25} & Scripted      & 30.5  & - \\
        \texttt{SPS-3}  & Spontaneous   & 5.5  & 30.0 \\
        \hline
        \multicolumn{2}{|c}{Total} & \textbf{36.0}  & \textbf{30.0}    \\ 
        \hline
        \multicolumn{4}{|c|}{} \\ 
        \hline
        \multicolumn{4}{|l|}{\textbf{out-of-domain (\texttt{OOD}) corpus}} \\ 
        \hline
        \texttt{Add\_data} & Radio         & 0.27  & - \\
        \hline
    \end{tabular}
    \caption{Datasets collected for ASR, expressed in hours.}
    \label{tab:dataset-corpus}
\end{table}
\section{ASR models for Puno Quechua (QXP)}
\label{sec:4-evaluation}

\paragraph{Datasets.} Two primary corpora were used: i) \texttt{SCS-25} with 30.5 validated hours of scripted speech; and ii) \texttt{SPS-3} with 5.5 validated hours of spontaneous speech (after excluding recordings longer than 30 seconds and adding 1 hour validated by a native speaker), supplemented by 30 hours of automatically generated silver transcriptions using \texttt{omniASR\_LLM\_7B} model. A small \texttt{OOD} corpus (\texttt{Add\_data}, ${\sim}$16 minutes) sourced from radio and social media, which was transcribed and validated manually by a native speaker, provides a third evaluation domain.

\paragraph{Foundation models.} 
We fine-tuned three architectures: 
(a) Whisper-base (74M parameters), an encoder-decoder Transformer trained on 680k hours of supervised multilingual speech. We fine-tuned setting the transcript prefix to Spanish. The unbalanced setting (\texttt{V}) used a learning rate (LR) of $5 \times 10^{-6}$ and the balanced setting (\texttt{V+S}) used a learning rate of $1 \times 10^{-5}$. Audio files longer than 30 seconds were excluded. 
% \sout{\textcolor{red}{The resulting model is available on HuggingFace.\footnote{\url{https://huggingface.co/rumiwarmi/whisper-base-qxp-finetuned}}}}
(b) wav2vec2-base (95M parameters), a self-supervised convolutional-Transformer model pre-trained on 960 hours of Librispeech. Both configurations were trained with LR: $1 \times 10^{-4}$, with stronger attention dropout (0.1) for the unbalanced corpus (\texttt{V}) to mitigate overfitting. Audio file longer than 20 seconds were excluded.  
and (c) XLS-R-300M (315M parameters), a multilingual wav2vec2 model pre-trained on 436k hours across 128 languages~\cite{babu2022xls}, making it the strongest starting point for low-resource languages with unusual phonological features such as ejectives and uvulars, and allophonic variations. A CTC projection head over a 46-character vocabulary (Puno Quechua Latin orthography) was added. Training runs for 20,000 updates with a tri-stage scheduler and LR: $5 \times 10^{-5}$. The encoder was frozen for the first 10,000 updates to prevent the randomly initialised CTC head from corrupting pre-trained representations before it stabilised. The best checkpoint was selected by validation WER.
% All models are available on HuggingFace. \footnote{\url{https://huggingface.co/QuechuaBase}} 

\paragraph{Continued Pre-Training (CPT).} For XLS-R-300M, we additionally performed CPT on the 65 hours of unlabelled Puno Quechua audio prior to fine-tuning. CPT adapts the model's acoustic representations to the target language without requiring transcriptions, and has demonstrated consistent gains in low-resource settings~\cite{dehaven2022improving,mutisya2026continued}.
Clips shorter than 1 second or longer than 15 seconds were excluded from training.
We train for 10,000 updates (LR: $1 \times 10^{-4}$, polynomial decay, 1,000-step warmup), selecting the best checkpoint by validation loss. The best checkpoint occurs at update 9,000 with a validation loss of 2.249. 
Two models were fine-tuned from the CPT checkpoint (\texttt{\small ft\_xlsr\_validated} and \texttt{\small ft\_xlsr\_silver}); using the identical protocol described above for fine-tuned XLS-R-300M.

\paragraph{Reference baselines.} We evaluated the omniASR model family~\cite{Keren25}, which combines a wav2vec2-style encoder with either CTC decoding (\texttt{CTC\_300M}, \texttt{CTC\_7B}; up to 6.5B parameters) or an LLM decoder (\texttt{LLM\_300M}, \texttt{LLM\_7B}; up to 7.8B parameters), and explicitly supports \texttt{qxp}. We also evaluated \texttt{MMS-1b-all} (1B parameters) by setting the language parameter to Cuzco Quechua (\texttt{quz}) for inference, the closest supported variety to Puno Quechua.\footnote{Both belong to the Collao linguistic subgroup and share a similar phoneme inventory and writing system.}
\section{Experiments and Results}
\label{sec:5-results}

\subsection{Baseline with off-the-shelf models}
As shown in Table~\ref{tab:omniasr-eval}, hybrid ASR-LLM models outperform CTC-only variants. The most balanced model across domains, omniASR LLM\_7B\_v2, achieves a 20.1\% mean WER. Notably, all omniASR models handle spontaneous speech more accurately than scripted speech and \texttt{OOD} (the CER is significantly better), maybe because the sentences in this corpus are very short (often between 3 and 5 words), and don't provide enough context.%, a pattern that warrants further investigation into vocabulary richness and speech rate differences among the corpora.

The MMS model, despite being trained on a different Quechua variety (Cuzco Quechua, \texttt{quz}), remains competitive, reaching 5.3\% CER on scripted speech. 
% \textcolor{red}{I'm sure that this is why its higlighted. but just to make sure we have to add OOD analysis} Added.

\begin{table}[t]
\def\arraystretch{1.2}
\centering
\scriptsize
\begin{tabular}{|l|c|c|c|c|c|c|}
\hline
 & \multicolumn{2}{c|}{\textbf{Scripted}} & \multicolumn{2}{c|}{\textbf{Spontaneous}} & \multicolumn{2}{c|}{\textbf{\texttt{OOD}}}\\ \hline
\textbf{Model} & \textbf{WER} & \textbf{CER} & \textbf{WER} & \textbf{CER} & \textbf{WER} & \textbf{CER} \\ \hline

\multicolumn{7}{|l|}{\textbf{omniASR}} \\ \hline
\texttt{CTC\_300M\_v2} & 47.8 & 10.3 & 29.0 & 4.4 &41.0	& 6.0\\ \hline
\texttt{CTC\_7B\_v2}   & 35.4 & 7.4 & 18.1 & 2.7 &34.5 & 5.7\\ \hline
\texttt{LLM\_300M\_v2} & \textbf{25.9} & 5.8 & 17.9 & 2.9 &24.4 & \textbf{3.9} \\ \hline
\texttt{LLM\_7B\_v2}   & 26.6 & 6.2 & \textbf{11.1} & \textbf{1.9} & \textbf{23.7} & 4.1\\ \hline

\multicolumn{7}{|l|}{\textbf{MMS}} \\ \hline
\texttt{mms-1b-all} & 35.0 & \textbf{5.3} & 36.4 & 6.5 & 38.0 & 6.2 \\ \hline
\end{tabular}
\caption{Performance of SOTA off-the-shelf models on 1,000 files samples for \texttt{qxp} (Scripted, Spontaneous) and on \texttt{OOD}.}
% \textcolor{red}{update title. I dont think this is 1000 files anymore? Or at least it isn't with OOD. THere also seems to be decimal point inconsistencies here}}
\label{tab:omniasr-eval}
\end{table}

\subsection{Results with fine-tuned models}
\paragraph{Training setup.}
Data were split 70/25/5 (train/dev/test). We compared two training configurations: 
\begin{itemize}
    \item \textbf{validated-only} (\texttt{V}, 36 hours) with 3× upsampling of spontaneous to compensate for class imbalance,
    \item \textbf{validated-plus-silver} (\texttt{V+S}, 66 hours) including silver spontaneous transcriptions.
\end{itemize}
Fine-tuning evaluation used three test sets: scripted (1.53 h), spontaneous (0.27 h), and \texttt{OOD} (0.27 h).
All the experiments were conducted on a 48GB L40S single GPU.

% \textcolor{red}{should we say "experiments were conducted on 48 GB L40S GPUs."}

\begin{table*}[!t]
\centering
\def\arraystretch{1.2}
\small
\begin{tabular}{ll|ll|ll|ll|ll|}
\cline{3-10}
&                  & \multicolumn{2}{c|}{\textbf{Scripted}} & \multicolumn{2}{c|}{\textbf{Spontaneous}} & \multicolumn{2}{c|}{\textbf{OOD}} & \multicolumn{2}{c|}{\textbf{Mean}} \\ \hline
\multicolumn{1}{|l|}{\textbf{Base model}} & \textbf{Dataset} & 
\multicolumn{1}{l|}{WER} & CER  & \multicolumn{1}{l|}{WER} & CER      & \multicolumn{1}{l|}{WER} & CER  & \multicolumn{1}{l|}{WER} & CER   \\ \hline
\multicolumn{1}{|l|}{\texttt{whisper-base}}& V
& \multicolumn{1}{l|}{8.57}  &  1.38
& \multicolumn{1}{l|}{26.2}  &  4.13       
& \multicolumn{1}{l|}{54.7}  &  10.8    
& \multicolumn{1}{l|}{29.8}  &  5.43     \\ \hline %MEAN
\multicolumn{1}{|l|}{\texttt{whisper-base}}        & V+S              
& \multicolumn{1}{l|}{3.81}  &  0.60
& \multicolumn{1}{l|}{17.1}  &  2.74        
& \multicolumn{1}{l|}{42.0}  &  7.77   
& \multicolumn{1}{l|}{21.0}  &  3.70     \\ \hline %MEAN
\multicolumn{1}{|l|}{\texttt{wav2vec2-base}}  & V    
& \multicolumn{1}{l|}{5.84}  & 0.77 
& \multicolumn{1}{l|}{21.6}  & 3.06         
& \multicolumn{1}{l|}{54.2}  & 10.3    
& \multicolumn{1}{l|}{27.2}  & 4.71      \\ \hline %MEAN
\multicolumn{1}{|l|}{\texttt{wav2vec2-base}}       & V+S              
& \multicolumn{1}{l|}{7.37}    & 0.96    
& \multicolumn{1}{l|}{13.9}    & 1.70     
& \multicolumn{1}{l|}{50.2}     & 9.45 
& \multicolumn{1}{l|}{23.8}     & 4.03      \\ \hline
\multicolumn{1}{|l|}{\texttt{xls-r-300m}}          & V   & \multicolumn{1}{l|}{2.06} & 0.30 & \multicolumn{1}{l|}{13.6} & 1.71 &
  \multicolumn{1}{l|}{35.5} & 6.03 & \multicolumn{1}{l|}{17.1} & 2.68 \\ \hline
  \multicolumn{1}{|l|}{\texttt{xls-r-300m}}  & V+S & \multicolumn{1}{l|}{4.36} & 0.57 & \multicolumn{1}{l|}{6.68}  & 0.81 &
  \multicolumn{1}{l|}{28.9} & 4.35 & \multicolumn{1}{l|}{13.3} & 1.91 \\ \hline
  \addlinespace[1em]\hline 
  \multicolumn{1}{|l|}{xls-r + CPT}    & V   & \multicolumn{1}{l|}{\textbf{1.19}} & \textbf{0.19} & \multicolumn{1}{l|}{13.6} & 1.73 &
  \multicolumn{1}{l|}{35.0} & 6.09 & \multicolumn{1}{l|}{16.6} & 2.67 \\ \hline
  \multicolumn{1}{|l|}{xls-r + CPT}    & V+S & \multicolumn{1}{l|}{2.11} & 0.30 & \multicolumn{1}{l|}{\textbf{3.15}} & \textbf{0.41} &
  \multicolumn{1}{l|}{\textbf{27.4}} & \textbf{4.55} & \multicolumn{1}{l|}{\textbf{10.9}} & \textbf{1.75} \\ \hline  
\end{tabular}
\caption{Performance of foundation models fine-tuned on validated corpus (V) and on complete corpus (V+S). WER and CER are expressed in \%.
% \textcolor{red}{there seem to be some decimal point inconsistencies}
}
\label{tab:results-finetuning}
\end{table*}

\paragraph{Fine-tuned and CPT results.} Table~\ref{tab:results-finetuning} reports WER and CER across all fine-tuned conditions. Three findings stand out. First, CPT yields consistent gains on scripted speech: XLS-R+CPT (\texttt{V}) achieves 1.19\% WER versus 2.06\% without CPT (a relative improvement of 42\%). Second, silver data is the decisive factor for spontaneous speech: XLS-R+CPT trained on \texttt{V+S} reduces spontaneous WER (13.6\% $\rightarrow$ 3.15\%, a relative reduction of 77\%); the same pattern holds without CPT (13.6\% to 6.68\%, a relative reduction of 51\%). 
Third, a clear \texttt{OOD} generalisation gap persists for all fine-tuned models: silver models achieve ${\sim}$35-54\% WER versus ${\sim}$27-50\% for validated-only models, but the omniASR LLM\_7B\_v2 still outperforms all fine-tuned systems on \texttt{OOD} (WER: 23.7\%), indicating that fine-tuning on in-domain data comes at some cost to out-of-domain robustness. 
% \textcolor{red}{I think this needs to be updated to reflect up to date numbers. Table 3 and including references to table 2 results.}

\section{Discussion}
\label{sec:6-discussion}

\paragraph{Silver data as the decisive factor for spontaneous speech.} Across all model architectures, including V+S silver transcriptions drastically reduces spontaneous WER. 
The effect is most noted for XLS-R+CPT (a relative reduction of 77\%) and is consistent even for Whisper-base (26.2\% $\rightarrow$ 17.1\%, a relative reduction of 34.7\%) and wav2vec2-base (21.6\% $\rightarrow$ 13.9\%, a relative reduction of 35.65\%). 
% \textcolor{red}{
This confirms that the low validation rate of spontaneous speech (14.7\%) is a bottleneck for improving ASR system's performance, and that automatically generated silver transcriptions, despite their lower quality, provide crucial coverage of speech variation.
% }

\paragraph{Effect of CPT on scripted speech.} CPT on unlabelled Puno Quechua audio provides consistent gains on scripted speech regardless of data configuration. 
% \textcolor{red}{
The relative improvement for XLS-R (\texttt{V}) (2.06\% $\rightarrow$ 1.19\%, a relative improvement of 42\%) demonstrates that adapting the pre-trained acoustic model to the target-language's phonology characteristics is valuable even when the same data are later used for fine-tuning~\cite{getman2024exploring}. The CPT benefit is somewhat diminished when silver data is added ( 4.36\% vs. 2.11\% without CPT), suggesting that silver data partially compensates for the lack of language-specific pre-training.
% }

\paragraph{OOD generalisation gap.} A clear generalisation gap exists for all fine-tuned models on out-of-domain data. 
% \textcolor{red}{
The omniASR \texttt{LLM\_7B\_v2} achieves the best \texttt{OOD} WER (23.7\%), outperforming the best fine-tuned system (XLS-R + CPT, \texttt{V+S}: 27.4\%). This suggests that task-specific fine-tuning on a narrow domain comes at the cost of robustness to unseen acoustic conditions and speaking styles. However, there is a substantial resource disparity involved. The omniASR \texttt{LLM\_7B\_v2} operates at ${\sim}$7.8B parameters and requires ${\sim}$30 GB of VRAM at inference\footnote{\url{https://huggingface.co/facebook/omniASR-LLM-300M}}, while our XLS-R+CPT competitive model operates at just 317M parameters and ${\sim}$2GB of VRAM, making it deployable on commodity hardware. 
% \textcolor{red}{we can say something like, even though we were 5X smaller, we had a similar performance to the leading LLM model. And beat other similar sized models by x amount} 
Closing the \texttt{OOD} gap for our model will therefore require not simply more training data, but exploration of lightweight strategies that preserve cross-lingual generalisation while remaining deployable in low-resource settings. 
% }

% \paragraph{Participatory design and dataset quality.} 
% The participatory design methodology produces a dataset that is qualitatively distinct from those generated by extractive or crowdsourcing approaches. By involving the Puno Quechua speech community, we collected speech covering domains directly relevant to the community's needs rather than Bible passages or generic read text. In terms of community engagement, participants were predominantly female (67.5\%), while speakers over 50 years of age were notably under-represented (7.9\%). Achieving a more balanced demographic distribution remains an objective for future campaigns.

% \paragraph{Dataset limitations.} One important limitation of the current corpus is the low validation rate for spontaneous speech (only 14.7\%). This reflects the challenge of transcribing spontaneous speech in a language with low written literacy rates among its speaker community.
% Addressing this gap is a near-term priority, through community-driven transcription campaigns, semi-supervised approaches using omniASR silver labels, or a hybrid strategy combining both.
\section{Conclusions and Future Work}
\label{sec:7-conclusion}

This paper makes three concrete contributions. 

\paragraph{Largest Puno Quechua ASR corpus.} We have constructed, to the best of our knowledge, the largest corpus ever prepared for ASR in a single Quechua variety. The corpus comprises 66 hours of recordings for scripted and spontaneous speech (including 36 hours of manually transcribed and validated data), supplemented by 30 hours of automatically transcribed silver spontaneous speech, and 0.27 hours of out-of-domain annotated data. The data were collected through a four-phase participatory design process involving 151 native speakers and released openly via Mozilla Data Collective under a \texttt{CC0-1.0} licence. The participatory methodology ensured that the corpus reflects domains directly relevant to the community.

\paragraph{Systematic ASR benchmark for Puno Quechua.} We establish the first variety-specific ASR benchmark for Puno Quechua, evaluating SOTA models (omniASR CTC and LLM variants up to 7B parameters; MMS-1b-all) and fine-tuned foundation models (Whisper-base, wav2vec2-base, XLS-R-300M, with and without CPT) across scripted, spontaneous, and out-of-domain test sets. Key findings are: (a) silver transcriptions are the decisive factor for spontaneous speech performance, reducing WER by up to 77\% relative; (b) continued pre-training on unlabelled Puno Quechua audio yields consistent gains on scripted speech; (c) omniASR models outperform all fine-tuned systems on out-of-domain data, revealing a generalisation gap that remains an open challenge.

\paragraph{Open release of fine-tuned models.} We release all fine-tuned model variants for Puno Quechua, including Whisper-base, wav2vec2-base, and XLS-R-300M under both \texttt{V} and \texttt{V+S} configurations, as well as the CPT checkpoint and the CPT-based fine-tuned models. The best-performing system, XLS-R-300M with CPT, fine-tuned on \texttt{V+S}, achieves 2.11\% WER and 0.30\% CER on scripted speech, and 3.15\% WER and 0.41\% CER on spontaneous.

Future research follows from these results. The enrichment of the corpus, in accordance with the quality requirements outlined above, must be continued. Another objective is to incorporate data representative of a wider range of domains, as we have begun to do through the annotation of data crawled from media sources. 
% Many experiments remain to be conducted to train robust ASR models. The richness of the annotated corpus could support the development or improvement of auxiliary tools such as diarization systems (for instance, pyannote\footnote{\url{https://github.com/pyannote/pyannote-audio}} currently performs poorly for Quechua) or Text-to-Speech (TTS) systems, which could in turn enable data augmentation.
With the ongoing goal of developing tools that genuinely meet users’ needs, we also aim to design more resource-efficient models (for example through quantization) that can be integrated into everyday applications, particularly mobile voice input systems.

\section*{Limitations}
Despite the contributions presented in this paper, some limitations should be acknowledged.
\paragraph{Corpus size.} Although the corpus is the largest for any single Quechua variety, only 14.7\% of the 35.3 recorded hours of spontaneous speech have been validated and transcribed, reflecting the difficulty of manual annotation in a community with low written literacy rates in Quechua. 
\paragraph{Out-Of-Domain.} All fine-tuned models exhibit a clear generalisation gap relative to the off-the-shelf models. Expanding the diversity of training and evaluation domains, e.g., radio, television, and social media, will be necessary to close this gap without sacrificing the parameter efficiency that makes our models deployable on commodity hardware.

\section*{Acknowledgements}
This research was supported by UK Research and Innovation (UKRI) Frontier Research Grant EP/Y031350/1 under the UK government’s funding guarantee for ERC Advanced Grants for the project entitled ``Towards Globally Equitable Language Technologies (EQUATE)''; by netidee Förderung (\url{www.netidee.at}); by SILICON Stanford (\url{silicon.stanford.edu}); and by the French National Research Agency and Ministry of Higher Education, Research and Innovation (MESR).

\bibliography{custom}

% \appendix

% \section{Example Appendix}
% \label{sec:appendix}

% This is an appendix.

\end{document}